\documentclass[11pt, a4paper, logo, onecolumn, internal, copyright, nonumbering]{deepmind}

\pdftrailerid{redacted}

\makeatletter
\renewcommand\bibentry[1]{\nocite{#1}{\frenchspacing\@nameuse{BR@r@#1\@extra@b@citeb}}}
\makeatother

\usepackage{kantlipsum, lipsum}
\usepackage{dsfont}
\usepackage{dm-colors}
\usepackage{multirow}
\usepackage[noend]{algorithm2e}
\newtheorem{theorem}{Theorem}
\usepackage{algorithm,algorithmic}

\usepackage[authoryear, sort&compress, round]{natbib} 

\graphicspath{{figures/}}

\title{Accelerating Large Language Model Decoding with Speculative Sampling}

\correspondingauthor{ccharlie@deepmind.com}

\author{Charlie Chen}
\author{Sebastian Borgeaud}
\author{Geoffrey Irving}
\author{Jean-Baptiste Lespiau}
\author{Laurent Sifre}
\author{John Jumper}
\affil{All authors from DeepMind}

\begin{abstract}
We present speculative sampling, an algorithm for accelerating transformer decoding by enabling the generation of multiple tokens from each transformer call. Our algorithm relies on the observation that the latency of parallel scoring of short continuations, generated by a faster but less powerful draft model, is comparable to that of sampling a single token from the larger target model. This is combined with a novel modified rejection sampling scheme which preserves the distribution of the target model within hardware numerics. We benchmark speculative sampling with Chinchilla, a 70 billion parameter language model, achieving a $2$--$2.5\times$ decoding speedup in a distributed setup, without compromising the sample quality or making modifications to the model itself.
\end{abstract}

\begin{document}
\maketitle
\section{Introduction}
Scaling transformer models to 500B+ parameters has led to large performance improvements on many natural language, computer vision and reinforcement learning tasks \citep{brown2020language, rae2021scaling, hoffmann2022chinchilla, chowdhery2022palm, arnab2021vivit, dosovitskiy2020image}. However, transformer decoding remains a highly costly and inefficient process in this regime.
\\\\
Transformer sampling is typically memory bandwidth bound \citep{mqattention}, so for a given set of hardware, the time to generate a single token in transformer models is proportional to a first order approximation to the size of parameters and the size of the transformer memory. The size of language models also necessitates serving with model parallelism -- adding communication overheads \citep{pope2022efficiently} and multiplying resource requirements. Since each new token depends on the past, many such transformer calls are required to sample a new sequence.
\\\\
We present an algorithm to accelerate transformer sampling for latency critical applications, which we call speculative sampling (SpS). This is achieved by:

\begin{enumerate}
	\item Generating a short draft of length $K$. This can be attained with either a parallel model \citep{blockwisedecoding} or by calling a faster, auto-regressive model $K$ times. We shall refer to this model as the \textbf{draft model}, and focus on the case where it is auto-regressive.
	\item Scoring the draft using the larger, more powerful model from we wish to sample from. We shall refer to this model as the \textbf{target model}.
	\item Using a modified rejection sampling scheme, accept a subset of the $K$ draft tokens from left to right, recovering the distribution of the target model in the process.
\end{enumerate}
Intuitively, there are often sequences where the next token might be ``obvious''. Therefore, if there is strong agreement between the draft and target model's distributions on a given token or sub-sequence of tokens, this setup permits the generation of \textit{multiple tokens each time the target model is called}.
\\\\
We show that the expected acceptance rate of draft tokens is sufficient to offset the overhead of the drafting process for large language models, resulting in an effective and practical method for reducing sampling latency without the need for modifying the target model or biasing the sample distribution. Depending on the evaluation domain, SpS leads to a $2$--$2.5\times$ speedup when sampling from Chinchilla \citep{hoffmann2022chinchilla}. Notably, the mean tokens per second with SpS often exceeds the idealised ceiling on auto-regressive sampling speed imposed by the memory bandwidth.
\section{Related Work}
There has been a substantial body of work focused on improving sampling latency of large transformers and other auto-regressive models.
\\\\
Since sampling performance is heavily coupled with the model size in memory, quantisation to \texttt{int8} or even \texttt{int4} \citep{dettmers2022llm, yao2022zeroquant} and distillation \citep{jiao-etal-2020-tinybert, sanh2019distilbert} of transformers are effective techniques for reducing sampling latency with little to no performance penalty. The observation that model size contributes less to the final performance than expected \citep{hoffmann2022chinchilla} has also encouraged smaller language models in general.
\\\\
During sampling, a cache of the keys and values is maintained for every attention layer, and could become a memory bandwidth bottleneck as the batch size increases. Methods such as multi-query attention  \citep{mqattention} aims to improve sampling performance by shrinking this cache. However these techniques are most effective at maximising throughout (at larger batch sizes) instead of latency, especially for larger models where the majority of the memory bandwidth budget is consumed by the parameters.
\\\\
Using a combination of the above techniques, in addition to a number of low-level optimisations to TPUs, \citet{pope2022efficiently} have greatly improved the serving latency and efficiency of PaLM 540B.
\\\\
There is an existing body of similar work exploiting the efficiency of transformers and sequence models operating in parallel. This includes block parallel sampling \citep{blockwisedecoding}, aggressive decoding \citep{aggressivedecoding}, in addition to some work in parallelizing autoregressive models in the image domain \citep{predictivesampling, parallelNES}. These methods have yet to be adapted to typical language model use-cases since they either only work with greedy sampling, bias the results or are focused on other modalities. Further, to our knowledge none of these techniques have been scaled to distributed setups, which is necessary for the most expensive decoders with the tens or hundreds of billions of parameters.
\\\\
Coincidentally, the work in this manuscript was undertaken concurrently and independently of the work on speculative decoding from \citet{speculativegoogle}. We focus more heavily the distributed serving setting for large models and offer some incremental optimisations, but otherwise the core underlying idea is the same.
\section{Auto-regressive Sampling}

Whilst transformers can be trained efficiently and in parallel on TPUs and GPUs, samples are typically drawn auto-regressively (See \autoref{alg:alg_ars}).
For most applications, auto-regressive sampling (ArS) is highly memory bandwidth bound and thus cannot make effective use of modern accelerator hardware \citep{mqattention}. A memory bound model call only generates a single token for every sequence in the batch, hence generating multiple tokens introduces a large amount of latency in any system which makes use of it. 
\\\\
This is especially problematic as the number of parameters in the model increases. Since all the model parameters need to pass through at least one accelerator chip, the model size divided by the total memory bandwidth across all chips gives us a hard ceiling on the maximum auto-regressive sampling speed. Larger models also require serving on multiple accelerators, introducing a further source of latency due to inter-device communication overheads.
\begin{algorithm}[!h]
\caption{Auto-regressive (ArS) with Auto-Regressive Models}
\label{alg:alg_ars}
\begin{algorithmic}
\STATE Given auto-regressive target model $q(.|.)$ and initial prompt sequence $x_1, …, x_t$ and target sequence length $T$.
\STATE Initialise $n \leftarrow t$. 
\WHILE{$n<T$} 
    \STATE Sample $x_{n+1} \sim q(x| x_1,   \dots, x_n)$
    \STATE $n \leftarrow n+1$
\ENDWHILE
\end{algorithmic}
\end{algorithm}

\begin{algorithm}[h]
 \caption{Speculative Sampling (SpS) with Auto-Regressive Target and Draft Models}\label{alg:alg_sps}
 
\begin{algorithmic}

  \SetAlgoLined
  \SetKwProg{Def}{def}{:}{}

  \DontPrintSemicolon

    \STATE Given lookahead $K$ and minimum target sequence length $T$.
    \STATE Given auto-regressive target model $q(.|.)$, and auto-regressive draft model $p(.|.)$, initial prompt sequence $x_0, …, x_t$.

\STATE Initialise $n \leftarrow t$.
 
\WHILE{$n<T$} 
 \FOR{$t = 1: K$}
  \STATE Sample draft auto-regressively $\tilde x_{t} \sim p(x| , x_1, \dots, x_n, 
\tilde x_1, \dots, \tilde x_{t-1})$

\ENDFOR
\STATE In parallel, compute $K+1$ sets of logits from drafts $\tilde x_1, \dots, \tilde x_{K}$ :\[
q(x| , x_1, \dots, x_n),\
q(x| , x_1, \dots, x_n, \tilde x_1),\ \dots,\ q(x| , x_1, \dots, x_n, \tilde x_1, \dots, \tilde x_K)
\]
\FOR{$t = 1: K$} 
    \STATE  Sample $r\sim U[0,1]$ from a uniform distribution. 
        \STATE \textbf{if} $r <\min\left(1, \frac{q(x|  x_1, \dots, x_{n+t-1})}{p(x|  x_1, \dots, x_{n+t-1})}\right)$, \textbf{then}
        \STATE \quad Set $x_{n+t}\leftarrow \tilde x_t$ and $n\leftarrow n+1$.
        \STATE \textbf{else}
        \STATE \quad sample $x_{n+t} \sim (q(x|  x_1, \dots, x_{n+t-1})-p(x|  x_1, \dots, x_{n+t-1}))_+$ and exit for loop.
        \STATE \textbf{end if}
\ENDFOR
\STATE If all tokens $x_{n+1}, \dots , x_{n+K}$ are accepted, sample extra token $x_{n+K+1} \sim q(x| , x_1, \dots, x_n, x_{n+K})$ and set $n\leftarrow n+1$.
\ENDWHILE
\end{algorithmic}
\end{algorithm}
\section{Speculative Sampling}
\subsection{Conditional Scoring}
For speculative sampling (See \autoref{alg:alg_sps}), we first make the observation that computing the logits of a short continuation of $K$ tokens in parallel has a very similar latency to that of sampling a single token.
We focus our attention on large transformers, sharded in the Megatron style \citep{shoeybi2019megatron}. For these models the majority of sampling time can be attributed to three components:
\begin{enumerate}
	\item \textbf{Linear Layers:} For small batch sizes,
	each linear layer only processes a small number of embeddings. This causes the dense matrix multiplies in the feed-forward layers, queries, keys, values computations and the final attention projection to become memory bound. For small $K$, this will continue to be memory bound and therefore take a similar amount of time. 
    \item \textbf{The Attention Mechanism:} The attention mechanism is also memory bound. During sampling, we maintain a cache of all the keys and values of the previous tokens in the sequence to avoid re-computation. These KV-caches are large, and accounts for the majority of the memory bandwidth utilisation for the attention mechanism. However, since the KV-cache size does not change as we increase $K$, there is little to no delta in this component.
    \item \textbf{All-reduces:} As models grow in size, its parameters need to be divided across multiple accelerators, leading to communication overheads. With Megatron, this manifests itself as an all-reduce after every feed-forward and attention layer.
    Since only the activations for a small number of tokens are transmitted, this operation is typically latency bound instead of throughput bound for both sampling and scoring (for small $K$).
    Again, this results in a similar amount of time spent in the two cases.
\end{enumerate}
Other sources of overhead may exist, depending on the exact transformer implementation. Therefore it is still possible that the choice of positioning encoding, decoding method (e.g. a sort might be required for nucleus sampling), hardware limitations etc. can introduce some deltas between scoring and sampling. However, if the conditions are met such that the above components dominate then scoring should not be significantly slower for small $K$.

\subsection{Modified Rejection Sampling}

We require a method to recover the distribution of the target model from samples from the draft model, and logits of said tokens from both models.
\\\\
To achieve this, we introduce the following rejection sampling scheme of the drafted tokens. Given a sequence of tokens $x_1, \dots, x_n$, and $K$ draft tokens $\tilde x_{n+1}, \dots,  \tilde x_{n+K}$ generated from $p(.|.)$, we accept $\tilde x_{n+1}$ with probability:
\[
\min\left(1, \frac{q(\tilde x_{n+1}|x_1, \dots, x_n)}{p( \tilde x_{n+1}|x_1,\dots, x_n)}\right)
\]
Where $q(\tilde x_{n+1}|x_1, \dots, x_n)$ and $p(\tilde x_{n+1}|x_1, \dots, x_n)$ are the probability of $\tilde x_{n+1}$ according to the target and draft models respectively, conditioned on the context so far.
\\\\
If the token is accepted, we set $x_{n+1} \leftarrow \tilde x_{n+1}$ and repeat the process for $\tilde x_{n+2}$ until either a token is rejected or all tokens have been accepted.
\\\\
If $\tilde x_{n+1}$ is rejected, we resample $x_{n+1}$ from the following distribution:
\[
x_{n+1} \sim (q(x|x_1, \dots, x_n) - p(x|x_1, \dots, x_n))_+
\]
Where $(.)_+$ denotes: \[(f(x))_+ = \frac{\max(0, f(x))}{\sum_x \max(0, f(x))}\]
By applying this sequentially, we recover the distribution of the target model for the accepted tokens (see proof in Theorem \ref{thm:recovers_dist}) within hardware numerics. Note that:
\begin{itemize} 
    \item At least one token will always be generated from a draft-accept loop -- if the first token is rejected, a valid token is resampled.
	\item Since the final token of the draft gives us the logits for the next token, if every drafted token is accepted, we can sample from it normally. This gives us a maximum of $K+1$ tokens per loop, over the naive implementation which would only return $K$ tokens.
\end{itemize}
With standard sampling methods such as nucleus, top-k sampling and adjusting temperature, we can modify the probabilities accordingly before applying this rejection sampling scheme. We have observed that the overall acceptance rate is robust to the exact parameters used.
\\\\
Because we do not interact with the body of the transformer itself, this method can be used in conjunction many other techniques for accelerating or optimising the memory use of sampling, such as quantisation and multi-query attention.
\section{Choice of Draft Models}
Since the acceptance criterion guarantees the distribution of the target model in our samples, we are free to choose the method for drafting a continuation as long as it exposes logits, and there is a high enough acceptance rate and/or low enough latency to break-even. There exist several approaches here:
\begin{itemize}
\item Incorporating draft generation into the target model, and train the model from the start. This is the strategy used by \citet{blockwisedecoding}, which adds multiple heads into the transformer to generate multiple tokens.
\item Using sequence level distillation \citep{sequenceleveldist} to generate a second model which predicts $K$ tokens in parallel. This strategy was employed by \citet{aggressivedecoding}.
\item Set a portion of the activations of the target model as an input to the draft model, and train the draft model with this input.
\end{itemize}
Although these methods will likely yield powerful drafts, they require a large number of data generated from the target model or changes to the target model. Sequence level distillation in particular would require a large compute budget. This makes them less practical for large scale applications.
\\\\
Whilst large language models produce better samples, intuitively there are "easier" tokens to predict for which smaller models may be sufficient. Therefore we may simply use a smaller version of the target language model as the draft and obtain high acceptance rates. This would also be convenient from an engineering and workflow perspective, since robust tooling for such models should already exist to train the target model in the first place.

\section{Results}
We train a 4 billion parameter draft model optimised for sampling latency on 16 TPU v4s -- the same hardware that is typically used to serve Chinchilla for research purposes. This model was trained with the same tokeniser and dataset as Chinchilla, with a slightly smaller width and with only 8 layers. The relatively few number of layers allows it to achieve a sampling speed of \texttt{1.8ms/token} compared to \texttt{14.1ms/token} for Chinchilla. For details, please refer to the hyperparameters in \autoref{tab:hypers}.
\\\\
For distributed setups it is insufficient to naively choose a small model as the draft, since different models have different optimal inference setups.
For example, it is typical to serve Chinchilla 70B on 16 TPU v4s (where it achieves the aforementioned \texttt{14.1ms/token}), whereas a chinchilla-optimal 7B achieves its lowest sampling latency on 4 TPU v4s (where it achieves \texttt{5ms/token}).
For smaller models, the additional memory bandwidth and flops are insufficient to offset the additional communication overhead between more devices -- serving a 7B on 16 TPUs actually \textit{increases} the latency.
This means the 7B would provide only a modest speedup if used as a draft with its optimal topology, and we will not make full utilisation of the hardware during drafting.
\\\\
We can sidestep this issue by training a wider model with a relatively few number of layers in order to minimise communication overhead. It has been observed that the performance of language models is relatively robust to changes in model aspect ratio \citep{levine2020depth}, so this allows us to serve a powerful draft model which can be sampled rapidly on the same hardware as the target model.
\subsection{Evaluation on XSum and HumanEval}

\begin{table}[t]
    \caption{\textbf{Chinchilla performance and speed on XSum and HumanEval with naive and speculative sampling at batch size 1 and $K=4$}. XSum was executed with nucleus parameter $p=0.8$, and HumanEval with $p=0.95$ and temperature $0.8$.}
    \centering
    \vspace{.5em}
    \begin{tabular}{lcccc}
    \toprule
    Sampling Method & Benchmark & Result & Mean Token Time & Speed Up \\
     \midrule
    \midrule
     ArS (Nucleus) & \multirow{2}{*}{XSum (ROUGE-2)} & 0.112 & 14.1ms/Token & $1\times$\\
     SpS (Nucleus) && 0.114 &7.52ms/Token  & $1.92\times$\\
     \midrule
     ArS (Greedy) & \multirow{2}{*}{XSum (ROUGE-2)} & 0.157 & 14.1ms/Token & $1\times$\\
     SpS (Greedy) && 0.156 &7.00ms/Token  & $2.01\times$\\
    \midrule
    \midrule
     ArS (Nucleus) &  \multirow{2}{*}{HumanEval (100 Shot)} & 45.1\%
     & 14.1ms/Token &  $1\times$\\
     SpS (Nucleus) & &47.0\%   & 5.73ms/Token & $2.46\times$ \\
    \bottomrule
    \end{tabular}
    \label{tab:xsum_human_eval_results}
\end{table}

We evaluate speculative sampling with Chinchilla on two tasks and summarize the results in \autoref{tab:xsum_human_eval_results}:
\begin{itemize}
\item The XSum \citep{narayanxsum} benchmark. This is a natural language summarisation task using a 1-shot prompt where we sample a total of 11,305 sequences with a maximum sequence length 128.
\item The 100-shot HumanEval task \citep{humaneval}. This is a code generation task involves the generation of 16,400 samples with a maximum sequence length of 512.
\end{itemize}
Even with greedy sampling, a single token deviating due to numerics could result in two sequences diverging wildly. Since pseudo-random seeds are processed differently between ArS and SpS, and because the different computation graphs lead to different numerics, we cannot not expect identical outputs. However, we expect the samples to come from the same distribution within numerics and we empirically verify this by evaluating these benchmarks.
\\\\
We run the tasks at batch size 1 with SpS and ArS. The time taken per SpS/ArS loop has low variance, and we can measure it directly from TPU profiles. To obtain the average speedup, standard deviations and other metrics, we log the amount of tokens generated for each speculative loop.
In \autoref{tab:xsum_human_eval_results} we show the performance on the XSum and HumanEval benchmarks for naive and speculative sampling with Chinchilla.
\\\\
We obtain a substantial speedup in both tasks, with HumanEval reaching speedups of almost $2.5\times$. Yet, we have parity in the benchmark metrics -- the underlying samples distribution is provably the same up to numerics, and this verifies that the draft model is not biasing the results empirically. In the case of HumanEval and greedy XSum, this speedup \textit{exceeded the theoretical memory bandwidth limit of the hardware} for autoregressive sampling (model size divided by the total memory bandwidth).

\subsection{Acceptance rate changes per domain}
It is apparent that the acceptance rate is dependent on the application and the decoding method. HumanEval achieves a significantly larger speedup — We hypothesize that this is due to a combination of code containing a lot of common sub-sequences (e.g. \texttt{for i in range(len(arr)):} would be relatively easy for a draft model to guess), is often decomposed into a smaller set of shorter tokens and the temperature value sharpening both the draft and target logits.
\begin{figure}[h]
	\centering
	\includegraphics[width=\columnwidth]{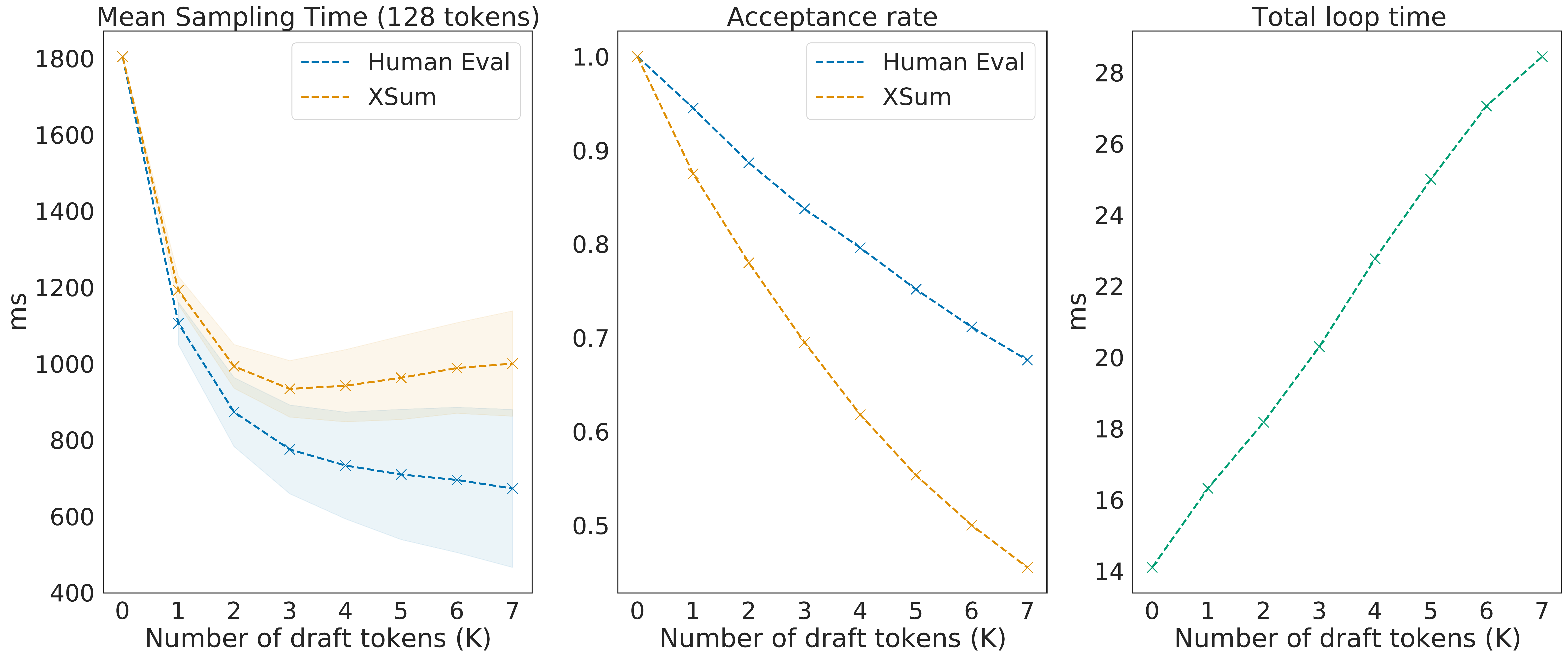}
	\caption{\textbf{Left:} The average time to generate 128 tokens, with standard deviation. Note that as $K$ increases, the overall speedup plateaus or even regresses, with XSum being optimal at $K=3$. The variance consistently increases with $K$. \textbf{Middle:} The average number of tokens accepted divided by $K+1$ -- this serves as a measure of the overall efficiency of the modified rejection scheme, which decreases with the lookahead. \textbf{Right:} Average time per loop increases approximately linearly with $K$ due to the increased number of model calls. Note that the gradient is slightly higher than the sampling speed of the draft model, due to additional overheads in nucleus decoding.}
	\label{fig:draft_len_tradeoff}
\end{figure}
\subsection{Trade off between longer drafts and more frequent scoring}
We visualise the trade-off of increasing $K$, the number of tokens sampled by the draft model in \autoref{fig:draft_len_tradeoff}.
As $K$ increases, we need fewer scoring calls from the large models to generate the same sequence length, potentially giving us a larger speedup.
However, the total loop time increases approximately linearly with the larger number of draft model calls and small increases in the scoring time.
The overall efficiency of the proportion of accepted tokens decreases as $K$ increases, since later tokens depend on the acceptance of previous tokens.
This results in the average speedup plateauing or even degrading with a larger $K$ (for example, XSum with nucleus's latency is minimised at $K=3$), depending on the domain.
\\\\
Further, even though larger values of $K$ may yield marginally greater mean speedups in certain circumstances, it also increases variance of the time to generate a full sequence. This could be problematic for settings where the P90, P99 latencies of concern.

\section{Conclusion}
In this work, we demonstrate a new algorithm and workflow for accelerating the decoding of language models.
Speculative sampling does not require making any modifications to the target language model's parameters or architecture, is provably lossless within numerics, scales well with the appropriate draft model and complements many existing techniques for reducing latency in the small batch size setting.
\\\\
We optimise and scale the technique to Chinchilla 70B using a draft model which was easy to train with existing infrastructure, demonstrating that it yields a large speedup across benchmark tasks and common decoding methods in the process. We verify that it is indeed lossless empirically in its downstream tasks. 

\bibliographystyle{abbrvnat}
\nobibliography*
\bibliography{template_refs}


\section*{Supplementary Materials}

\subsection*{Author Contributions}

\begin{itemize}
    \item \textbf{Initial proposal:} Charlie Chen, John Jumper and Geoffrey Irving
    \item \textbf{Initial Implementation, Optimisation and Scaling:} Charlie Chen
    \item \textbf{Modified Rejection Sampling Scheme:} John Jumper
    \item \textbf{Engineering Improvements:}  Jean-Baptiste Lespiau and Charlie Chen
    \item \textbf{Experiments:} Charlie Chen, Sebastian Borgeaud and Laurent Sifre
    \item \textbf{Draft of Manuscript:} Charlie Chen and Sebastian Borgeaud
    \item \textbf{Manuscript Feedback:} Laurent Sifre, Geoffrey Irving and John Jumper

\end{itemize}

\subsection*{Acknowledgements}
We'd like to thank Oriol Vinyals and Koray Kavukcuoglu for your kind advice and leadership.
We'd also like to thank Evan Senter for your additional feedback on the manuscript and Amelia Glaese for your support in navigating the publishing process.
Finally, we'd like to thank Blake Hechtman, Berkin Ilbeyi for your valuable advice on XLA and Nikolai Grigoriev for our discussions on the various tricks that can be applied to the transformer architecture.

\subsection*{Hyperparams}
\begin{table}[h]
    \caption{\textbf{Hyperparameters} for the draft model}
    \centering
    \vspace{.5em}
    \begin{tabular}{lcccc}
    \toprule
    Model & $d_\text{model}$ & Heads & Layers & Params\\
     \midrule
     Target (Chinchilla) & 8192 & 64 & 80 & 70B \\

     Draft & 6144 & 48 & 8 & 4B\\
    \bottomrule
    \end{tabular}
    \label{tab:hypers}
\end{table}
\subsection*{Proofs}
\begin{theorem}[Modified Rejection Sampling recovers the target distribution]\label{thm:recovers_dist}
Given discrete distributions $q$, $p$ and a single draft sample $\tilde x \sim p$, let $X$ be the final resulting sample. For $X=x$ to be true, we must either sample $\tilde x  = x$ and then accept it, or resample it after $\tilde x$ (of any value) is rejected. Hence:

\[\mathbb{P}(X=x) \]\[ =   \mathbb{P}( \tilde x  = x)  \mathbb{P}( \tilde x \text{ accepted}|  \tilde x  = x) +  \mathbb{P}(\tilde x \text{ rejected}) \mathbb{P}(X = x|  \tilde x \text{ rejected}) \]
For the first term, we apply the acceptance rule: 
\[ \mathbb{P}( \tilde x  = x)  \mathbb{P}( \tilde x \text{ accepted}|  \tilde x  = x) \]\[ =  p(x) \min\left(1, \frac{q(x)}{p(x)}\right) \]\[ =  \min\left(p(x), q(x)\right)\]
For the second conditional term, we apply the resampling rule:
\[ \mathbb{P}(X = x|  \tilde x \text{ rejected})  = (q(x) - p(x))_+\]
Where $(.)_+$ denotes: \[(f(x))_+ = \frac{\max(0, f(x))}{\sum_x \max(0, f(x))}\]
Finally, we calculate the probability of rejection:
\[ \mathbb{P}(\tilde x \text{ rejected}) = 1- \mathbb{P}(\tilde x \text{ accepted}) \]\[= 1 - \sum_{x'}\mathbb{P}(X=x',  \tilde x \text{ accepted})\]
\[ = 1 - \sum_{x'} \min(p(x'), q(x'))\]\[ =\sum_{x'} \max(0, q(x')-p(x')) \]
\[ = \sum_{x'}q(x') - \min(p(x'), q(x'))\]\[ =\sum_{x'} \max(0, q(x')-p(x')) \]
This is equal to the denominator of $(q(x) - p(x))_+$, so:
\[ \mathbb{P}(\tilde x \text{ rejected}) \mathbb{P}(X = x|  \tilde x \text{ rejected}) = \max(0, q(x) - p(x))\]
Hence:
\[\mathbb{P}(X=x) \]\[= \min(p(x), q(x))+ \max(0, q(x) - p(x)) \]\[= q(x)\]
and we have recovered the desired target.

\end{theorem}

\end{document}